\documentclass[12pt,draftcls,onecolumn]{IEEEtran}

\usepackage{epsfig}
\usepackage{graphicx}
\usepackage{amsmath}
\usepackage{amssymb}
\usepackage{amsfonts}
\usepackage{subfigure}
\usepackage{txfonts}
\usepackage{comment}
\usepackage{color}
\usepackage[ruled,linesnumbered]{algorithm2e}
\usepackage[nospace,noadjust]{cite}

\usepackage[pagebackref=true,
breaklinks=true,
letterpaper=true,
dvipdfm,
colorlinks=true,
linkcolor=red,
hyperindex,
citecolor=green,
pdfstartview=FitH,
plainpages=false,
bookmarks=true
]{hyperref}

\ifCLASSINFOpdf
\else
\fi

\begin{document}
%
\title{A Survey of Recent View-based 3D Model Retrieval Methods}
\author{Qiong Liu
\thanks{Qiong Liu is with Department of Electronics \& Information Engineering, Huazhong University of Science \& Technology, Wuhan, China.}
}

\markboth{Research Report}%
{Liu \MakeLowercase{\textit{et al.}}: A Survey of Recent View-based 3D Model Retrieval Methods}

\maketitle

\begin{abstract}
Extensive research efforts have been dedicated to 3D model retrieval in recent decades. Recently, view-based methods have attracted much research attention due to the high
discriminative property of multi-views for 3D object representation. In this report, we summarize the view-based 3D model methods and provide the further research trends. This paper focuses on the scheme for matching between multiple views of 3D models and the application of bag-of-visual-words method in 3D model retrieval. For matching between multiple views, the many-to-many matching, probabilistic matching and semi-supervised learning methods are introduced. For bag-of-visual-words application in 3D model retrieval, we first briefly review the bag-of-visual-words works on multimedia and computer vision tasks, where the visual dictionary has been detailed introduced. Then a series of 3D model retrieval methods by using bag-of-visual-words description are surveyed in this paper. At last, we summarize the further research content in view-based 3D model retrieval.
\end{abstract}


\IEEEpeerreviewmaketitle

\section{Introduction}
\label{sec:introduction}
With the fast development of internet technology, computer hardware, and software, 3D models \cite{Bimbo2007,Tangelder2008} have been widely
used in many applications, such as computer graphics, computer vision, CAD and medical imaging. Effectively and efficiently retrieve 3D model retrieval \cite{Yang2007,Leng2010,Tao2008,Paquet2000, Vranic2003,Gao2010MMBNT,Xiao2011a,Leibe2003,Xiao2011b} has attracted much research attention these days.

3D model retrieval methods can be divided into two categories: model-based methods and view-based methods. Early works are mainly model-based methods, in which low-level feature-based methods \cite{Leng2008,Tangelder2003,Mademlis2009} (e.g. the geometric moment \cite{Paquet2000a}, surface distribution \cite{Osada2002},
volumetric descriptor \cite{Tangelder2003,Funkhouser2003}, and surface geometry \cite{Ip2002,Johnson1999,Makadia2010,Pajarola2004,Paquet2000b}) or
high-level structure-based methods \cite{Leng2009} are employed. Due to the requirement of 3D models, these methods are limited in the practical applications.

Extensive research efforts have been dedicated to view-based 3D model retrieval methods \cite{Li2009,Li2010} because of the high discriminative property of multi-views for 3D object representation \cite{Ohbuchi2008b,Ohbuchi2008a}. Many view-based 3D object retrieval methods (e.g. Light Field Descriptor (LFD)~\cite{Chen2003}, Elevation Descriptor (ED)~\cite{Shih2007}, Bag-of-Visual-Features (BoVF) \cite{Ohbuchi2008b}, and Compact Multi-View Descriptor (CMVD)~\cite{Daras2010}) have been proposed these years. This is due to the fact that view-based methods \cite{Ansary2007TMM} are with the highly discriminative property for object representation and visual analysis \cite{Wang2007MMW,Gao2008TCE}  also plays an important role in multimedia applications.

The advantages of the view-based method are twofold.
\begin{enumerate}
    \item It does not require the explicit virtual model information, which makes the method robust to real practical applications.
    \item Image processing has been investigated for many decades. The view-based 3D model analysis methods can be benefited from existing image processing technologies.
\end{enumerate}

In this technical report, we review related works and recent trends in view-based 3D model retrieval, especial focus on the multiple-view matching scheme and the application of existing bag-of-words methods in 3D model retrieval.

\section{Multiple View Matching}
For view-based 3D model retrieval, each 3D model is represented by a group of 2D views. How to perform multiple view matching  is the key topic in view-based 3D model retrieval task. In this section, we briefly review existing multiple view matching methods.

Chen et al. first proposed the Light Field Descriptor (LFD) \cite{Chen2003}. LFDs are computed from 10 silhouettes obtained from the vertices of a dodecahedron over a
hemisphere. This image set described the spatial structure information from different views. In LFD, Zernike moments and Fourier descriptors of the 3D model were employed as the features of each image. This method found the best match between two LFDs as the similarity between two 3D models.

Ansary et al. \cite{Ansary2007} introduced an Adaptive Views Clustering (AVC) method . In AVC, there are 320 initial views  which are captured and representative views are optimally selected by adaptive view clustering with Bayesian information criteria. A probabilistic method is then employed to calculate the similarity between two 3D models, and those objects with high probability are selected as the retrieval results. There are two parameters in the method, which are used to modulate the probabilities of objects and views, respectively.

Daras et al. proposed a Compact Multi-View Descriptor (CMVD) \cite{Daras2010} method, in which 18 characteristic views of each 3D model are first selected through 18 vertices of the corresponding bounding 32-hedron. In CMVD, both the binary images and the depth images are taken to represent the views. Then the comparison between 3D models was based on the feature matching between selected views using 2D features, such as 2D Polar-Fourier Transform, 2D Zernike Moments, and 2D Krawtchouk Moments. For the query object, the testing object rotated and found the best matched direction for the query object. The minimal sum of
distance from the selected rotation direction was calculated to measure the distance between two objects.

In \cite{Mahmoudi2002}, 7 representative views from three principal and four secondary directions were acquired to index objects. The contour-based feature was extracted
for each view for multi-view matching. In \cite{Gao2010MM1}, query views were re-weighted using the relevance feedback information by multi-bipartite graph reinforcement model. In this method, the weights of query views were generated using the information propagation from the labelled retrieval results. Dealing with the view set of each 3D model, an incremental representative view selection method \cite{Gao2011TMM} has been proposed. In this method, the representative views for the query model is selected by using the user relevance feedback, and a distance metric is learnt for each selected views. Gao et al. \cite{Gao2012INS} proposed to update the weights for representative views by using pseudo-relevance feedback information, where a graph-based learning process is conducted to renew the query views' weights.

Some methods employed the generated view to represent 3D models. Panoramic object representation for accurate model attributing (PANORAMA) \cite{Papadakis2010} employed panoramic views to capture the position of the model's surface information as well as its orientation as the 3D model descriptor. The panoramic view of
a 3D model was obtained by projecting the 3D model to the lateral surface of a cylinder aligned with one of the object's three principal axes and centered at the centroid of the object.

Gao et al. proposed a Spatial Structure Circular Descriptor (SSCD) \cite{Gao2010PR}, which  can preserve the global spatial structure of 3D models, and it was invariant to rotation and scaling. All spatial information of 3D model can be represented by an SSCD which included several SSCD images. In SSCD, a minimal bounding sphere of the 3D model was computed, and all points on the 3D model surface were projected to the bounding sphere. Attribute values are provided with each point to represent the surface spatial information. The bounding sphere was further projected to a circular region of a plane. It can preserve the spatial structure of the original 3D model. This circular image was employed by each SSCD image to describe the surface information of a 3D model. Each spatial part of a 3D model is represented by one part of the SSCD individually. Histogram information was employed by SSCD as the feature of SSCDs to compare two 3D models.

Shih et al. proposed an Elevation Descriptor (ED) \cite{Shih2007}. ED extracts six range views to describe the original 3D model from its bounding box. These views contain the altitude information of the 3D model from six directions. 3D models are compared based on the matching of EDs. To match two groups of EDs, the minimal distance with order is calculated to measure the distance between the two 3D models.

Many exiting distance measures have also been investigated in view-based 3D model retrieval, e.g., the Hausdorf distance \cite{Atallah1983,Dubuisson1994}, the Earth Mover's Distance \cite{Rubner2000} and the bipartite graph matching \cite{Gao2008CIVR}. Gao et al. \cite{Gao2011SPIC} proposed a bipartite graph matching-based 3D model comparison method. In this method,  the weighted bipartite graph matching (WBGM), is employed for comparison between two 3D models. In the view-based 3D model retrieval, each 3D model is represented by a set of 2D views. Representative views are first selected and the corresponding initial weights are provided and further updated using the relationship among representative views. The weighted bipartite graph is built with these selected 2D views, and the proportional max-weighted
bipartite matching method \cite{Gao2008CIVR} is employed to find the best match in the weighted bipartite graph. Wen et al. \cite{Wen2012} further extended the bipartite graph matching method in 3D model matching, in which the constructed bipartite graph was first partitioned into subsets, and the matching was conducted in each subset.

Gao et al. \cite{Gao2012TIP1} proposed a probabilistic framework to compare two 3D models. In this method,  for every query object, we first cluster all query views to generate the view clusters, which are then used to build the query gaussian models. For more accurate 3D object comparison, a positive matching model and a negative matching model are individually trained using positive matched samples and negative matched samples, respectively. The CCFV model is generated on the basis of the query gaussian models by combining the positive matching model and the negative matching model.

Semi-supervised learning (graph-based) \cite{Zhou2004} has been applied on both multimedia indexing and retrieval tasks \cite{Wang2012ACM,Wang2010TMMb}, and has shown its superiority on investigating both the labeled and unlabeled data.  Wang et al. \cite{Wang2009a} employed graph-based semi-supervised learning on multi-concept detection. A multi-graph based semi-supervised learning method is introduced in \cite{Wang2009b} to fuse information from different sources by using a multi-graph framework. Towards a diverse and relevant  search of social images, the graph-based semi-supervised learning has been employed in \cite{Wang2010TMM} to learn the relevance scores for image to the concept. A semi-supervised kernel density estimation method is introduced in \cite{Wang2009CVIU} to annotate videos. In this method, the kernel density is estimated under the semi-supervised learning framework, and the optimal video annotation is achieved by the learning in the framework. Tang et al. \cite{Tang2009TSMCB} introduced to construct correlative linear neighborhood connection in the graph structure to learn the optimal relevance scores from video annotation. Semi-supervised learning on the hypergraph structure \cite{Zhou2007} has been investigated in many multimedia and computer vision tasks, e.g., image search \cite{Huang2009,Gao2011MM,Huang2010,Gao2012TIP2}. A hypergraph-based 3D object representation method is presented in \cite{Wong1989}, which constructs the hypergraph by using the correlation among different surface boundary segments of an object in the CAD system. Yu et al. \cite{Yu2012TIP} introduced a hypergraph learning method for image classification. In this method, the relevance scores among images are learned associated with the weights of constructed hyperedges. A hypergraph learning-based social image method is proposed in \cite{Gao2012TIP2}. A class-specific hypergraph (CSHG) \cite{Xia2008a} is proposed to integrate local SIFT and global geometric constraints for object recognition, where the hypergraph is employed to model a specific category of objects with multiple appearance instances.  Gao et al. \cite{Gao2012TIP3} introduced to formulate the relationship among different 3D models in a hypergraph structure. Each view of one 3D model is treated as the feature of the 3D model. The view clustering is first performed to generate view groups, and each view group generates one hyperedge, in which the 3D models with views in that view group are connected by this hyperedge. The semi-supervised learning on this constructed hypergraph structure is conducted to generate the relevance score among 3D models, which can be further used for 3D model retrieval.

\section{Bag-of-Visual-Words and Its Application in 3D Model Retrieval}
Generally,  the local features \cite{Lowe04}\cite{Mikolajczyk05} extracted from images are quantized into a set of visual words, where a visual word dictionary is created to generate an indexing file. Each image can be described by using  a Bag-of-Words histogram. This Bag-of-Words representation offers sufficient robustness against photographing variances in occlusions, viewpoints, illuminations, scales and backgrounds. In the application of Bag-of-Words method, the employed visual vocabulary for visual words plays an important role in the whole algorithm. In this section, we first introduce the existing visual vocabulary training method in detail, and following we provide the application of the Bag-of-Words method in 3D model retrieval.

\subsection{Visual Vocabulary Training in the Bag-of-Word method}
Typically speaking, building visual vocabulary usually resorts to unsupervised vector quantization\cite{Nister06}\cite{Philbin07}\cite{JiIJCV12}\cite{JiCVPR09}\cite{JiCVPR12b}\cite{JiIJCV12}, which subdivides the local feature space into discrete regions each corresponds to a visual word.
An image is represented as a Bag-of-Words (BoW) histogram, where each word bin counts how many local features of this image fall in the corresponding feature space partition of this word.
To this end, many vector quantization schemes are proposed to build visual vocabulary,
such as K-means \cite{Sivic03}, Hierarchical K-means (Vocabulary Tree) \cite{Nister06}, Approximate K-means \cite{Philbin07},
and their variances \cite{Jurie05} \cite{JiMM11}\cite{Yang07}\cite{Philbin07}\cite{JiCVPR12b} \cite{Jegou10}\cite{JiCVPR09}\cite{JiCVPR12c}.
Meanwhile, hashing local features into a discrete set of bins and indexed subsequently is an alternative choice, for which methods like Locality Sensitive Hashing (LSH) \cite{Gionis99}, Kernalized LSH \cite{Grauman09}, Spectral Hashing \cite{Weiss08} and its variances \cite{JiCVPR12a}\cite{JiTOMCCAP12b} are also exploited in the literature.
The visual word uncertainty and ambiguity are also investigated in \cite{Jegou08}\cite{Philbin07}\cite{Jiang07}\cite{Gemert09}, using methods such as
Hamming Embedding \cite{Jegou08}, Soft Assignments \cite{Philbin07} and Kernelized Codebook \cite{Gemert09}.
Some other related directions include optimizing the initial inputs of visual vocabulary construction, such as learning a better local descriptor detector as in \cite{JiTIST12}, coming up with a better similarity metric, such as learning an optimal hashing based distance matching as in \cite{JiPR11} for human action search  \cite{JiMM08}, incorporating Bayesian reasoning into the similarity calculation \cite{JiIJICIC08}, image annotation, landmark search \cite{JiMM09}\cite{JiMM09c}, text detection, and distributed visual search \cite{JiTMM12}.

Stepping forward from unsupervised vector quantization, semantic or category labels are also exploited \cite{Moosmann06}\cite{JiCVPR10}\cite{Mairal08}\cite{Lazebnik09}\cite{JiTIP12a} to supervise the vocabulary construction, which learns the vocabulary to be more suitable for the subsequent classifier training, \emph{e.g.}, the images in the same category are more likely to produce similar BoW histograms and vice versa.
The quantization issues in visual vocabulary are recently also well addressed to fit the city-scale landmark search scenario, such as the works in \cite{JiCVPR09}\cite{JiIEEEMM11}.
And more recently, the compression of visual vocabulary model has also received a wide variety of research interests.

\subsection{Application of the Bag-of-Words Method in 3D Model Retrieval}

To apply the bag-of-visual-words method to view-based 3D model retrieval, generally the SIFT features are extracted from all selected views, and a visual word dictionary is learnt. A bag-of-visual-words description is generated for 3D model representation. To perform 3D model matching, KL divergence or other distance metric can be employed.
 By using the visual words description, the detailed feature of each view can be represented well. Based on this description, it is disciminative for 3D model representation.

Furuya and Ohbuchi  have proposed a series of Bag-of-Visual-Words-based 3D model retrieval methods. In \cite{Furuya2008a}, they first involved the bag-of-visual-words method to view-based 3D model retrieval. They proposed to extract SIFT features for views of 3D models. In this method, each 3D model is rendered into a group of depth images, and the SIFT features are extracted from these images. This method uses the bag-of-features approach to integrate the local features into a feature vector for each model. Then the matching of these two feature vectors determines the distance between the two 3D models. Ohbuchi et al. \cite{Ohbuchi2008c} further proposed to employ Kullback-Leibler divergence (KLD) to calculate the distance between two bag-of-visual-feature based 3D models.  Ohbuchi and Shimizu \cite{Ohbuchi2008b} employed the semi-supervised manifold learning method for model class recognition.  The proposed method projects the original feature space onto a lower dimensional manifold. Then the relevance feedback information is employed to capture the semantic class information by using the manifold ranking algorithm. Though the bag-of-visual-feature description is effective on 3D model retrieval, the computational cost is high. Ohbuchi and Furuya \cite{Ohbuchi2008a} further accelerated the method by using a Graphics Processing Unit. Furuya and Ohbuchi \cite{Ohbuchi2009a} proposed to employ dense sampling to extract feature points in the views of 3D models,  and then the SIFT feature is extracted from each point. These visual features are further clustered into groups to generate the visual vocabulary. Then the feature histogram is generated to calculate the 3D model distance. This method has been further extended \cite{Ohbuchi2009a,Ohbuchi2010b} to deal with large scale data. A distance metric learning method \cite{Ohbuchi2010a} is proposed to learn the distance metric for matching of 3D models.

Osada et al. \cite{Osada2008a} employed the bag-of-visual-feature method to SHREC'08 CAD model track task.
A bag-of-region-words method \cite{Gao2010MM2} is introduced to extract visual features in the region level. This method first gridly selects points in each image and the local SIFT features are extracted for these points. Then each feature is encoded into a visual word with a pre-trained visual vocabulary. In this step, each view is split into a set of regions, and each region is represented by a bag-of-visual-words feature vector. All the obtained regions are further grouped into clusters based on the bag-of-visual-words feature, and one feature is chosen as the representative one from each cluster with corresponding weight.  The Earth Movers Distance is used to measure the distance between two 3D models. Endoh et al. \cite{Endoh2012} introduced  to conduct learning on the manifold structure of 3D models by using clustering-based training sample reduction. Kawamura et al. \cite{Kawamura2012} further employed the geometrical feature to improve the feature-based method.

\section{Discussion and Future Work}
View-based 3D model retrieval has been investigated in recent years, and many methods have been proposed recently. The key point for view-based 3D model retrieval lies in the matching of multiple views. Many works employ many-to-many matching schemes, e.g., EMD and bipartite graph matching, to measure the distance between two 3D models. Other works employ the probabilistic framework to estimate the similarity between 3D models. With the widely application of bag-of-visual-words method in multimedia and computer vision tasks, it has been employed in view-based 3D model retrieval. We have detailed introduced recent works by using bag-of-visual-words.

Though there are significant progress for view-based 3D model retrieval, there are still many problems which require further investigation.
\begin{enumerate}
  \item The description method for multiple views. Though many works have focused on view representation, e.g., by using Zernike Moments and Bag-of-Visual-Words, precisely multiple views description is still important for 3D model retrieval.
  \item Estimation the relationship among 3D models by multiple views. Most of existing methods employ many-to-many matching methods to calculate the distance between 3D models. A few methods use the probabilistic method to formulate the relationship among 3D models. As each 3D model is described by a group of views, the relationship among 3D models is complex than the relationship between just two images.
\end{enumerate}

\ifCLASSOPTIONcaptionsoff
  \newpage
\fi

\bibliographystyle{IEEEtran}

\end{document}